\pdfoutput=1
\documentclass[11pt]{article}

\usepackage[final]{acl}

\usepackage{times} % (Alternatives like txfonts or newtx are also acceptable.)
\usepackage{latexsym}

\usepackage{algorithm}
\usepackage{algorithmic}
\usepackage{xcolor}
\usepackage{tikz}

\usepackage{xspace}
\newcommand{\tool}{{\bf\textsc APE}\xspace}

\newcommand{\ignore}[1]{}
\usepackage{enumitem}
\usepackage{listings}
\usepackage[T1]{fontenc}
\usepackage[utf8]{inputenc}

\usepackage{microtype}

\usepackage{inconsolata}

\title{APE: Active Learning-based Tooling for Finding Informative Few-shot Examples 
for LLM-based Entity Matching}

\author{
    Kun Qian\thanks{Work done while working at Apple}\and
    Yisi Sang\and Farima Fatahi Bayat\thanks{Work done while interning at Apple}\and Anton Belyi\and Xianqi Chu\\
    {\bf Yash Govind} \and {\bf Samira Khorshidi} \and {\bf Rahul Khot } \and {\bf Katherine Luna} \and {\bf Azadeh Nikfarjam } \\
    {\bf Xiaoguang Qi } \and {\bf Fei Wu\thanks{Work done while interning at Apple}  } \and {\bf Xianhan Zhang } \and {\bf Yunyao Li\thanks{Work done while working at Apple}}\\
    \texttt{kunqian, yisi\_sang, f\_fatahibayat, a\_belyy, xchu23, yash\_govind, samiraa} \\
    \texttt{ r\_khot, kluna, anikfarjam, xiaoguang\_qi, fwu7, xianhan\_zhang, yunyaoli@apple.com} \\
}

\begin{document}
\maketitle
\begin{abstract}
Prompt engineering is an iterative procedure often requiring extensive manual effort to formulate suitable instructions for effectively directing large language models (LLMs) in specific tasks. Incorporating few-shot examples is a vital and effective approach to providing LLMs with precise instructions, leading to improved LLM performance. Nonetheless, identifying the most informative demonstrations for LLMs is labor-intensive, frequently entailing sifting through an extensive search space.
In this demonstration, we showcase a human-in-the-loop tool called \tool (Active Prompt Engineering) designed for refining prompts through active learning. Drawing inspiration from active learning, \tool iteratively selects the most ambiguous examples for human feedback, which will be transformed into few-shot examples within the prompt. Demo recording can be found with the submission or be viewed at {\color{blue}\url{https://youtu.be/OwQ6MQx53-Y}}.
\end{abstract}

\section{Introduction}
Prompt engineering typically serves as the initial step when developing LLM-based applications because it is a relatively fast process and requires fewer technical skills than fine-tuning.
Prompt engineering involves crafting and optimization of instructions provided to LLMs. These prompts need to be carefully designed to direct the behavior of LLMs towards performing specific tasks or generating desired outcomes \cite{prompt-based-learning}. 
While LLMs (e.g., ChatGPT and GPT-4) show impressive capabilities for zero-shot tasks without prior training, their performance can be further enhanced by integrating clear and informative few-shot demonstrations alongside the prompts \cite{white2023prompt}. These demonstrations not only guide the LLMs but also provide examples that contribute to more accurate and contextually relevant outputs, especially for ambiguous cases. 

Prompt engineering is a dynamic and iterative process that typically consists of the following stages: (1) \textit{Task Description}: clearly outline the intended task for LLMs, (2) \textit{Few-shot Demonstration}: provide a small number of concrete and helpful demonstrations to illustrate the precise semantics of the task, (3) \textit{Task Input and Completion Request}: present the actual task input and request an LLM completion. For all three steps, minor prompt rephrasing is typically needed, but this task is relatively light and does not require many iterations. However, choosing informative few-shot demonstrations can be a labor-intensive and time-consuming process due to the large search space of the problem. For instance, to identify only 3 demonstration examples out of 100 examples, there are 970,200 (i.e., 100$\times$99$\times$98) different combinations, a daunting manual task. 

\looseness=-1 Identifying representative and ambiguous examples to enhance the performance of machine learning models is a well-established subject within the active learning community.  We can view the few-shot example identification as an active learning problem, where the goal is to find the most informative examples to be included in the prompt to help improve LLMs' performance. Recently, \cite{active-prompt} proposed the idea of using various active learning sampling strategy to identify few-shot examples prompt engineering. Our work follows the same direction with the main focus being building an interactive tool (with an intuitive user interface) that identifies the most informative few-shot examples through simple human interaction. 
\begin{figure}[ht] 
    \centering
    \resizebox{\columnwidth}{!}{% Resize the TikZ picture to fit the column width
        \input{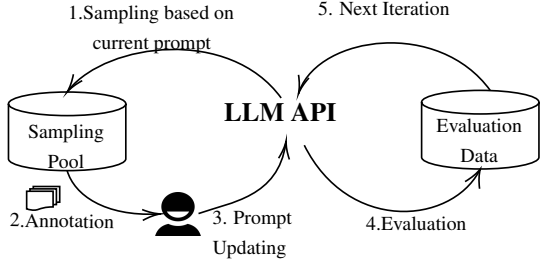}% Include the TikZ picture
    }
    \vspace{-3mm}
    \caption{System Overview}
    \label{fig:overview}
    \vspace{-3mm}
\end{figure}

%\begin{figure}[ht!] 
%    \scriptsize
%    \centering
%    \input{system_overview}
%    \includegraphics[width=0.9\columnwidth]{system_overview}
%    \vspace{-3mm}
%    \caption{System Overview}
%    \label{fig:overview}
%    \vspace{-3mm}
%\end{figure}
In this paper, we present \tool (\textbf{A}ctive \textbf{P}rompt \textbf{E}ngineering), an intuitive and intelligent prompt engineering tool that iteratively identifies the most informative and ambiguous examples for which a given LLM will likely make a mistake, and then provide them in the prompt as few-shot examples after seeking human annotation. Unlike \cite{active-prompt}, which focuses on the backend algorithm services, our goal is to hide the technical details by a carefully designed graph user interface so that we can have a usable tool that truly harnesses the power of active learning. 
%Inspired by active learning, a human-in-the-loop learning paradigm that systematically and iteratively identifies uncertain examples to improve a machine learning model via human feedback, we can also view few-shot example identification as an active learning problem, where the goal is to find the most informative examples to be included in the prompt to help improve LLMs' performance.  
 %Concretely, instead of manually and potentially randomly crafting few-shot examples, we can use active learning to identify the most informative and uncertain examples as few-shot examples. In fact, a recent work named Active-Prompt has proposed a similar idea \cite{active-prompt}. In this paper, we present \tool (\textbf{A}ctive \textbf{P}rompt \textbf{E}ngineering). This intuitive, intelligent, prompt engineering tool identifies the most informative few-shot examples through simple human interaction. \tool uses an active-learning-style process to iteratively identify examples for which LLMs will likely make a mistake and then provide them in the prompt as few-shot examples after seeking human annotations. Our main goal is to build a tool with a friendly interface that truly harnesses the power of active learning. 
\section{Methodology}
The main goal of \tool is to identify a handful of informative few-shot examples that can boost an LLM's performance. As an active learning tool, it follows the iterative procedure outlined in Figure \ref{fig:overview}, involving interaction with both a human user and the LLM API for prompt engineering.

The best way to understand \tool end to end is to watch the video demo of the tool (see the link in the abstract). At a high level, in each iteration, we start with sampling informative examples based on the prompt of the current iteration, which includes applying a user-configured sampling strategy to let the LLM choose the ambiguous examples from the user-provided sampling pool. Next, users annotate the selected examples, potentially including explanations for Chain-of-Thought-style prompting. These newly annotated examples are then used to update the prompt. Lastly, the new prompt is evaluated against evaluation data to report its performance.

The core of the active learning process is the sampling strategy; for simplicity, we will use entity matching, a classic binary classification task, to illustrate the sampling methodology behind \tool. Given a set $P=\{p_1, \dots, p_m\}$ of entity pairs, where $p_i$ consists of a pair $\langle e_1^i, e_2^i\rangle$ of entities, the task of entity matching is to learn a binary classifier $f: \langle e_1^i, e_2^i \rangle \rightarrow \{0, 1\}$.  In this case, the binary classifier is the LLM in consideration, and the behavior of the classifier is dynamically controlled by the prompts created by \tool. 
In our demo video, we used the DBLP-Scholar dataset sampled from \cite{kopcke2010evaluation} to illustrate \tool. 

\subsection{Active Sampling Strategy}
The core of \tool is to find the most informative examples for human annotation to boost the performance of the LLMs. LLMs can be considered excellent student models that can learn effectively from examples. Inspired by active learning, we proposed identifying examples that LLMs are uncertain about to be used as few-shot examples for in-context learning. 
While both task-specific sampling strategies and task-agnostic sampling strategies can be integrated with \tool, due to limited space, we focus on the task-agnostic approaches because they allow \tool to be easily applied to a wide range of problems. In this demo, we introduce two task-agnostic strategies: (1) random-based sampling and (2) self-consistency-based sampling. 

\smallskip
\noindent \textbf{Random-based}. We randomly select $k$ examples (no replacement) from the sampling pool in each active iteration. Random sampling is simple and fast, and it would work reliably well for many simple tasks. However, for more sophisticated tasks where zero-shot LLMs do not perform well, the chance that random sampling would find informative examples to boost LLMs' performance is low.

\smallskip
\noindent \textbf{Self-consistency-based}. To overcome the issue of random sampling, we support self-consistency-based sampling, a strategy inspired by self-consistency \cite{wang2023selfconsistency}.  The core idea is to either run multiple different prompts or the same prompt multiple times in the style of Chain-of-Thought \cite{wei2023chainofthought}, allowing the model to generate the final answers with multiple reasoning paths. The consistent answers (e.g., the majority answer) are then chosen as the final answer. A similar idea, known as query-by-committee (QBC)\cite{qbc}, has been heavily used in active learning to identify uncertain examples \cite{settles2009active}. QBC works by training a committee of $k$ slightly different classifiers, e.g., five deep-learning-based classifiers with different architectures, and then let the committee make inferences over the same examples. The disagreement ratio of the committee is used as a proxy to quantify the uncertainty of the examples. The examples with high disagreement ratios are then sent for human annotation. 

Our self-consistency-based strategy follows the same idea. Concretely, when selecting examples from the sampling pool, for every entity pair $\langle e_1, e_2 \rangle$, we run the same prompt $m$ times, where $m$ is a hyperparameter that is usually a small number (in our case, 3). However, each run of the prompt would use a different temperature $t$, where $t$ gradually grows from 0 to 1 depending on the number of runs. For instance, if $m=3$, then the three runs of the prompt would have temperatures: $0, 0.5, 1.0$, respectively. Varying the temperature is a way to control the creativity and consistency of LLMs, and we use it to build a committee of slightly different LLMs for uncertain example sampling. Specifically for our entity matching demo scenarios, we collect the $m$ binary labels for a given entity pair $p$, we then compute the label distributions of the $m$ predictions.  We denote the ratio of positive labels as $R^+(p)$ (i.e., $\frac{\textup{\em \# positive labels}}{m}$), and obviously the ratio of negative labels would be $1-R^+(p)$. With that, we can then compute the label distribution entropy $H(p)$ as follows:
\[
-R^+(p)\log{R^+(p)} - (1-R^+(p))\log{(1-R^+(p))}
\]
the entropy can be viewed as a proxy for uncertainty, and the higher the entropy value, the higher the uncertainty. We then select the examples with the top-$k$ entropy (breaking tie arbitrarily) for human annotations. The annotated examples will be included as new few-shot examples. Note that varying temperatures is for sampling mode only, we set the temperature to zero during prompt evaluation. 

\smallskip
\noindent\textbf{Incremental or Fixed Sampling}.
We offer both incremental sampling and fixed sampling. Incremental sampling accumulates examples labeled in each iteration to form the final few-shot demonstrations. In contrast, fixed sampling selects a predetermined number of examples iteration without accumulating them to create the final prompt.

\smallskip
\noindent\textbf{Human Annotation}. By default, we only ask the annotator for the ground truth of a selected example, but for self-consistency-based, we also ask for an explanation of the label provided. Both settings are user-configurable.

\section{Concluding Remarks}
Due to limited space, we focus on the tooling aspect of \tool in this demo paper, and are currently working on a research paper that will provide a comprehensive description of the system design, theoretical foundation underlying this optimization problem, and experimental evaluations.

% propose a theoretical framework that can 
% comprehensively describe the system design as well as experimental evaluations. 
% \section{Demo System \& Demo Scenarios}
% Figure \ref{fig:overview} gives an overview of \tool. To set up \tool, the following three inputs are required: (1) An instructable LM for which we need to use \tool to identify optimal prompts for certain task, e.g., GPT-turbo-3.5 used in this demo; (2) a jsonl file containing unlabeled examples designated for sampling few-shot demonstrations; (3) a jsonl file containing evaluation examples accompanied by human annotations, which will be used to evaluate the performance of actively learned prompts. The user also needs to choose sampling strategy and what kind of annotation will be provided. Please see the video demo recording ({\color{blue}\url{https://youtu.be/OwQ6MQx53-Y}}) for more details.  
%This short demo paper focuses on the tooling aspect of the active prompt engineering work, we are planning a research paper to comprehensively describe the system design as well as experimental evaluations. 

\bibliography{custom}

\begin{thebibliography}{8}
\expandafter\ifx\csname natexlab\endcsname\relax\def\natexlab#1{#1}\fi

\bibitem[{Diao et~al.(2023)Diao, Wang, Lin, and Zhang}]{active-prompt}
Shizhe Diao, Pengcheng Wang, Yong Lin, and Tong Zhang. 2023.
\newblock \href {http://arxiv.org/abs/2302.12246} {Active prompting with chain-of-thought for large language models}.

\bibitem[{K{\"o}pcke et~al.(2010)K{\"o}pcke, Thor, and Rahm}]{kopcke2010evaluation}
Hanna K{\"o}pcke, Andreas Thor, and Erhard Rahm. 2010.
\newblock Evaluation of entity resolution approaches on real-world match problems.
\newblock \emph{Proceedings of the VLDB Endowment}, 3(1-2):484--493.

\bibitem[{Liu et~al.(2023)Liu, Yuan, Fu, Jiang, Hayashi, and Neubig}]{prompt-based-learning}
Pengfei Liu, Weizhe Yuan, Jinlan Fu, Zhengbao Jiang, Hiroaki Hayashi, and Graham Neubig. 2023.
\newblock \href {https://doi.org/10.1145/3560815} {Pre-train, prompt, and predict: A systematic survey of prompting methods in natural language processing}.
\newblock \emph{ACM Comput. Surv.}, 55(9).

\bibitem[{Settles(2009)}]{settles2009active}
Burr Settles. 2009.
\newblock Active learning literature survey.

\bibitem[{Seung et~al.(1992)Seung, Opper, and Sompolinsky}]{qbc}
H.~S. Seung, M.~Opper, and H.~Sompolinsky. 1992.
\newblock \href {https://doi.org/10.1145/130385.130417} {Query by committee}.
\newblock In \emph{Proceedings of the Fifth Annual Workshop on Computational Learning Theory}, COLT '92, page 287–294, New York, NY, USA. Association for Computing Machinery.

\bibitem[{Wang et~al.(2023)Wang, Wei, Schuurmans, Le, Chi, Narang, Chowdhery, and Zhou}]{wang2023selfconsistency}
Xuezhi Wang, Jason Wei, Dale Schuurmans, Quoc~V Le, Ed~H. Chi, Sharan Narang, Aakanksha Chowdhery, and Denny Zhou. 2023.
\newblock \href {https://openreview.net/forum?id=1PL1NIMMrw} {Self-consistency improves chain of thought reasoning in language models}.
\newblock In \emph{The Eleventh International Conference on Learning Representations}.

\bibitem[{Wei et~al.(2023)Wei, Wang, Schuurmans, Bosma, Ichter, Xia, Chi, Le, and Zhou}]{wei2023chainofthought}
Jason Wei, Xuezhi Wang, Dale Schuurmans, Maarten Bosma, Brian Ichter, Fei Xia, Ed~Chi, Quoc Le, and Denny Zhou. 2023.
\newblock \href {http://arxiv.org/abs/2201.11903} {Chain-of-thought prompting elicits reasoning in large language models}.

\bibitem[{White et~al.(2023)White, Fu, Hays, Sandborn, Olea, Gilbert, Elnashar, Spencer-Smith, and Schmidt}]{white2023prompt}
Jules White, Quchen Fu, Sam Hays, Michael Sandborn, Carlos Olea, Henry Gilbert, Ashraf Elnashar, Jesse Spencer-Smith, and Douglas~C. Schmidt. 2023.
\newblock \href {http://arxiv.org/abs/2302.11382} {A prompt pattern catalog to enhance prompt engineering with chatgpt}.

\end{thebibliography}

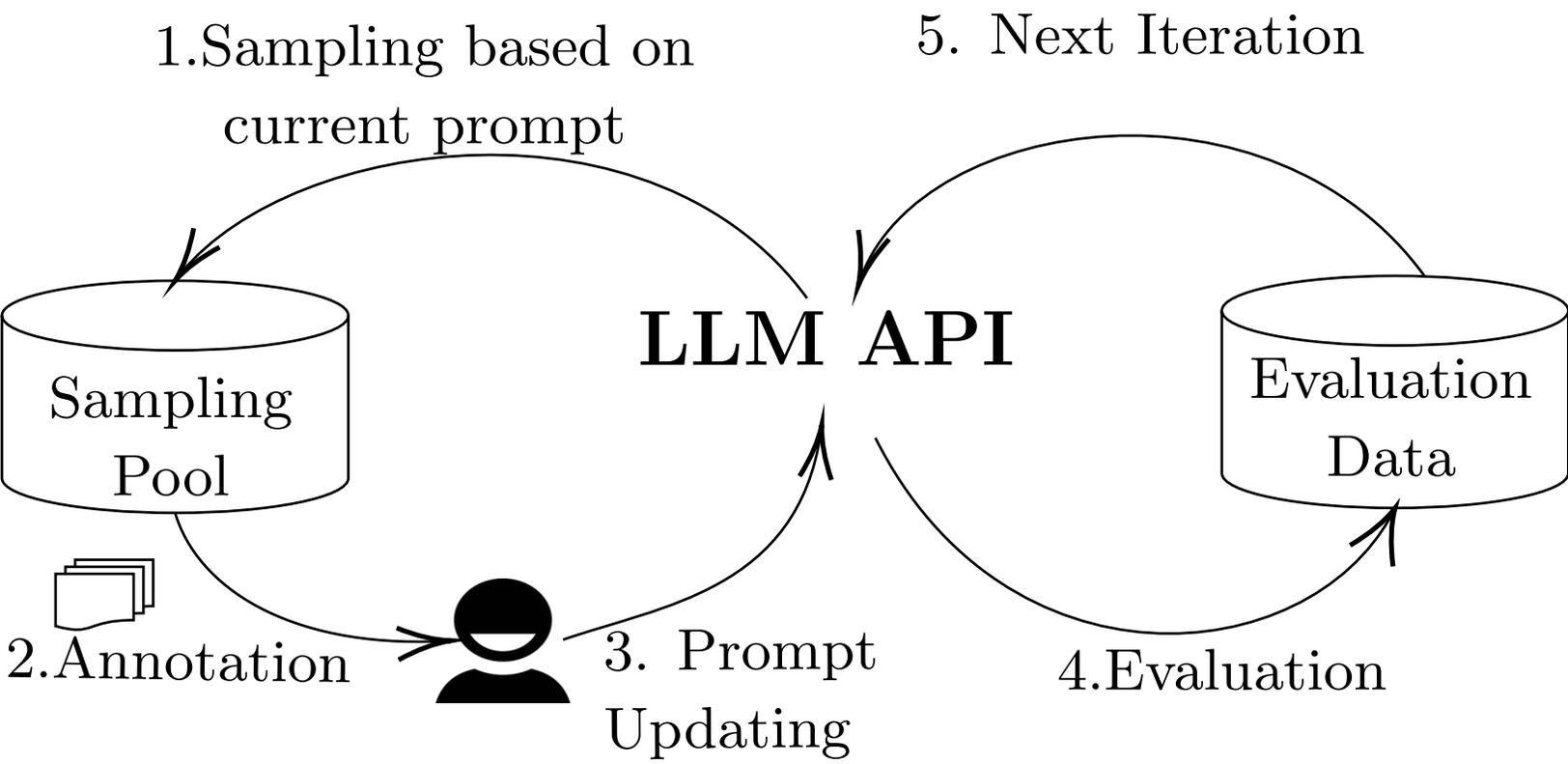
\end{document}